\documentclass{article}

\usepackage[preprint, nonatbib]{neurips_2020}

\usepackage[utf8]{inputenc} 
\usepackage[T1]{fontenc}    
\usepackage{hyperref}       
\usepackage{url}            
\usepackage{booktabs}       
\usepackage{amsfonts}       
\usepackage{nicefrac}       
\usepackage{microtype}      
\usepackage{graphicx}
\usepackage{amsmath}
\graphicspath{ {./figures/} }

\title{A Meta-Bayesian Model of Intentional Visual Search}

\author{%
	Maell Cullen\\
	Department of Engineering Mathematics\\
	University of Bristol\\
	\texttt{maell.cullen@bristol.ac.uk} \\
	\And
	M. Berk Mirza \\
	Department of Neuroimaging \\
	King's College London \\
	\texttt{muammer.mirza@kcl.ac.uk} \\
	\And
	Jonathan Monney \\
	Department of Neuroimaging \\
	King's College London \\
	\texttt{jonathan.monney@kcl.ac.uk} \\
	\And
	Rosalyn Moran \\
	Department of Neuroimaging \\
	King's College London \\
	\texttt{rosalyn.moran@kcl.ac.uk} \\
}

\begin{document}
	\maketitle
	
	\begin{abstract}
		We propose a computational model of visual search that incorporates Bayesian interpretations of the neural mechanisms that underlie categorical perception and saccade planning. To enable meaningful comparisons between simulated and human behaviours, we employ a gaze-contingent paradigm that required participants to classify occluded MNIST digits through a window that followed their gaze. The conditional independencies imposed by a separation of time scales in this task are embodied by constraints on the hierarchical structure of our model; planning and decision making are cast as a partially observable Markov Decision Process while proprioceptive and exteroceptive signals are integrated by a dynamic model that facilitates approximate inference on visual information and its latent causes. Our model is able to recapitulate human behavioural metrics such as classification accuracy while retaining a high degree of interpretability, which we demonstrate by recovering subject-specific parameters from observed human behaviour. 
	\end{abstract}
	
	\section{Introduction}
	The way we see the world is contingent upon the way we move our eyes, behind every eye movement is an inferential process that determines what to place within the central $2^{\circ}$ of the visual field that is processed by 50\% of our primary visual cortex \cite{mason1991central}. Consciously or not, whatever it is that we see, it is usually the case that we sought to see it. This presumption is supported by the pre-motor theory of attention \cite{rizzolatti1987reorienting} which suggests the neural mechanisms underlying the planning and execution of eye movements are closely linked with those responsible for attentional modulation. From this we may presume that by analysing eye movements we may gain insight into the inferential processes guiding their deployment.  
	
	In the setting of behavioural experiments, observed behaviours are often contextualised by objective task contingencies or rules, yet less frequently considered are the subjective perceptual representations that enable an individual to recognise these contingencies. In this paper we propose a computational model of visual search that facilitates inference on an individual's behaviour and their subjective perceptual inferences, a problem that has previously been described as one of meta-Bayesian inference \cite{daunizeau2010observing}, i.e., making inferences about inferences. We evaluate our model with a gaze-contingent (moving-window) paradigm in which human participants were asked to classify handwritten digits from the MNIST dataset \cite{lecun2010mnist}. In this task the contents of the visual scene were occluded beyond a window that followed participant's gaze. 
	
	\begin{figure}[hbt!]
		\centering
		\includegraphics[width=12cm]{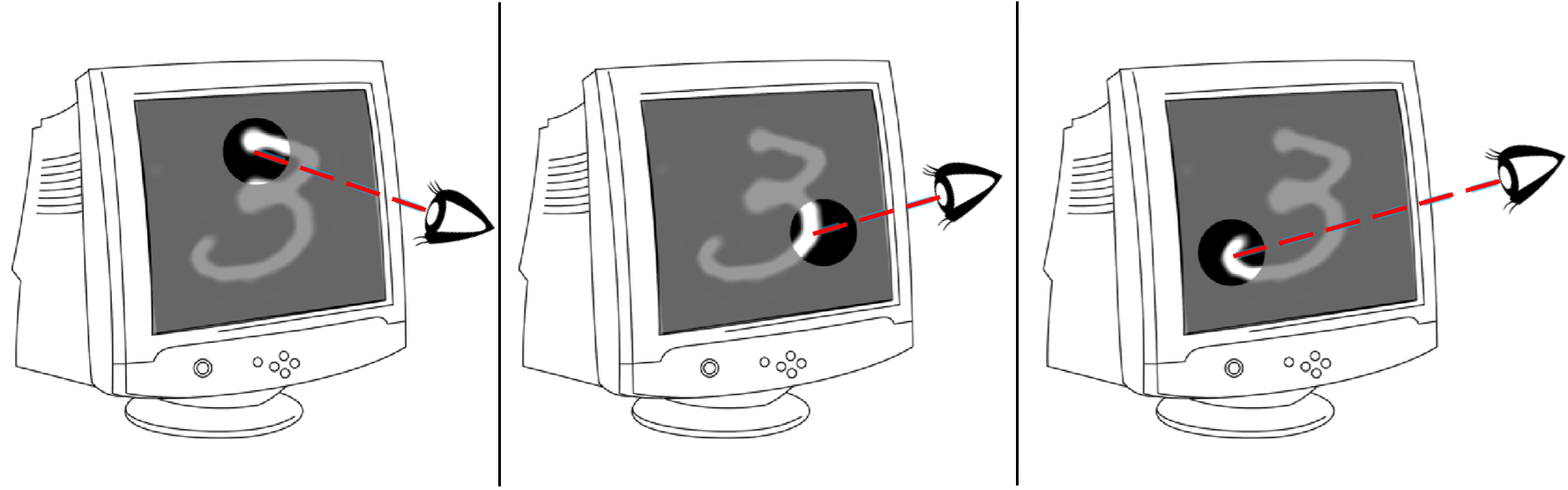}
		\caption{\textbf{Gaze-contingent paradigm.} Human participants were asked to classify 100 digits as quickly and as accurately as possible. Answers were reported by pressing a button and focusing their gaze on the corresponding choice location. Note that the grey areas of the mask are made transparent for illustrative purposes only.}
	\end{figure}
	
	\section{Motivation}
	
	Perhaps the most influential model of visual attention is Itti’s implementation \cite{itti1998model} of Koch \& Ullman’s  computational theory of saliency \cite{koch1987shifts} in the primate visual system. However, this and other models of attentional selection that omit the influence of an agent’s internal states and intentions will be challenged by the complexity and scope of many behavioural tasks \cite{Yarbus1967}, in part due to an inability to dynamically reattribute salience with respect to new information. The role of top-down attentional modulation has featured more\ prominently in recent models \cite{einha2008task}\cite{ballard2009modelling} of eye movements, and the concept of ‘salience’ has given some way to that of ‘priority’. The deployment of spatial attention and salience attribution has been attributed to ‘priority maps’ encoded by interactions between lateral interparietal area  \cite{gottlieb1998representation} \cite{bisley2010attention}, frontal eye fields \cite{schall2002neural}\cite{thompson2005neuronal} and the superior colliculus \cite{krauzlis2013superior}. 
	
	Neural representations of visual salience are not serially processed one fixation at a time \cite{shen2014predictive} but through a process wherein the lateral interparietal area and frontal eye fields select saccade targets in advance of saccades and project to intermediate layers of the superior colliculus to influence lower motor pathways. We reflect this process in the implementation of feature or priority maps in the state-space of our model; while evaluating potential saccade locations the agent will be influenced by the conditional feature probabilities encoded by these maps, allowing it to allocate attention over regions of visual space that are likely to confirm (or refute) hypotheses about the identity of the digit. In our formulation we assume that the priority maps for each digit class are computed a-priori, yet their influence is effectively dynamic, as the agent relies upon ascending messages from the dynamic model to weigh their relative utility. We view this analogously with participants having knowledge about the general form of each digit and utilising this information to locate visually salient visual features. 
	
	Due to the retinocentric organisation of neurons in the visual cortex, representations of the visual field may switch or change dramatically between fixations. To account for this, the motor cortex generates a copy of its output in the form of ‘corollary discharge’ \cite{sperry1950neural}. These signals are sent to visual areas to inform predictions of reafferent visual feedback, i.e, in distinguishing retinal displacement from movement in the external environment \cite{sommer2008brain} and of proprioceptive signals from the eye muscles \cite{wang2007proprioceptive}. In our model these signals, induced during policy selection in the Markov Decision Process (MDP), impose empirical priors on the output of a dynamic model that generates predictions about the outcomes of action \cite{friston2017active}. Empirical proprioceptive priors here are simply target locations in 2D visual space while the empirical exteroceptive priors are defined over the latent space of a variational autoencoder, which acts as a proxy for the forward (generative) models in the sensory cortex that compute visual predictions.
	
	\section{Formulation}
	All inference schemes based on probability density functions can be reformulated as optimisation problems under a variational formulation of Bayes rule. In brief, this involves approximating the true posterior with a variational (proposal) density that can be optimised with respect to observed data \cite{beal2003variational}. Optimisation in this context corresponds to minimizing variational free energy, a lower bound \cite{feynman1972statistical} on the approximate log-evidence of a model. This technique is commonly used in machine learning for approximating intractable posterior densities with neural networks \cite{dayan1995helmholtz}\cite{kingma2014semi}\cite{mnih2014neural}. It is also central to the Free Energy Principle, a mathematical formulation of how adaptive systems resist a natural tendency to disorder \cite{friston2010free}\cite{friston2019free} and Active inference, a neurobiologically influenced process theory of how the neuronal mechanisms of action and perception are unified by this objective \cite{friston2017active}. In the following section we show how perception and behaviour (eye movements) can be formulated as inference on the hidden states of the world, and how these processes can be simulated by optimising the variational free energy of a generative model of the environment. 
	
	We first construct our visual foraging task as a partially observable Markov decision process with categorical task outcomes. Under this model beliefs are expressed as the joint density of observations, hidden states, policies and precision:
	\[ P(\tilde{o}, \tilde{s}, \pi,\gamma) = P(\pi|\gamma)P(\gamma) \prod_{t=1}^{T}P(o_t|s_t)  P(s_t|s_{t-1,}\pi)\]
	
	Where a likelihood matrix $P\left(o_{t}=i| s_{t}=j\right)=A_{ij}$ defines the probability of an outcome o under every combination of hidden states $ s $ and the transition matrix $P\left(s_{t}=i| s_{t-1}=j,\pi\right)={B\left(u\right)}_{ij}$ defines the probabilistic mapping from the hidden states at the current time step to the hidden states at the next time step under some action $ u $. Under this model, outcomes $ o $ are determined only by the current state $ s_{t} $ and beliefs about state transitions are determined by policies, where each policy $\pi$ comprises a series of actions $u=\pi\left(t\right)$. The mapping between policies and hidden states is influenced by the agent’s prior preferences, or the epistemic value of each outcome: 
	\[ P(o):=\sigma(C) \]
	
	The precision parameter $ \gamma $ determines an agent’s confidence in its decisions, or expected uncertainty, defined as an inverse-gamma distribution with precision $ \beta $:
	
	\[ P\left(\gamma\right)=\Gamma\left(1,\beta=\frac{1}{\gamma}\right) \]
	
	The state-space of the MDP is comprised of 3 primary hidden state factors, $ \mathit{digit} $, $ \mathit{where} $ and $ \mathit{report} $. The $ \mathit{digit} $ factor defines the possible target classes.  The $ \mathit{where} $ factor defines regions in visual space to which the agent can saccade. There are 49 such locations arranged in a  7x7  grid and an additional location that the agent must saccade to before making a decision. The $ \mathit{report} $ factor defines control states that the agent may invoke to either report the target class or remain undecided. In all trials the undecided state persists through the first transition, giving the agent enough time to forage for information before having to report its decision. Finally, a variable number of \textit{feature} factors define visual features such as contrast or orientation. While the number of factors is determined a-priori by a saliency-map algorithm, the possible states within each feature factor represent the presence of the corresponding feature, for example 1 and 5 may represent ‘None’ and ‘Strong’ contrast, respectively. We found that \cite{itti1998model}\cite{garcia2012saliency} and \cite{harel2007graph} produced suitable class-contingent saliency maps.
	
	The model considers 3 outcome modalities, \textit{digit}, \textit{where} and \textit{feedback}. The \textit{digit} and \textit{where} outcomes are mapped directly to their corresponding hidden state factors while the third modality provides the agent with \textit{feedback} in the form 3 possible outcomes, correct, incorrect or undecided. If the invoked control state from the \textit{report} factor is aligned with the target class the model will observe correct feedback, which is associated with high utility, otherwise it will receive incorrect feedback which is associated with low utility. Note that the causal structure of the hidden states precludes direct influence of extrinsic reward on the instantaneous belief updates that occur within the subordinate (continuous) level.
	
	Having defined a generative model, the agent’s approximate posterior Q can be computed by inverting the generative model, allowing the agent to form expectations about the hidden states:
	\[ \begin{gathered} 
	Q(\tilde{s},\pi)=Q(\pi)\prod_{t=1}^{T}{Q(s_{t}|\pi)} 
	\end{gathered} \]
	
	In our gaze-contingent task, visual observations depend upon sequences of saccades requiring the generative model to entertain expectations under different policies, or sequences of actions. The equation above states that the approximate posterior can be factorised by taking the product of the marginal state and policy distributions over time, assuming that control states are approximately independent from one another at each time step. The variational free energy may now be defined with respect to this factorised distribution:
	\[ \begin{gathered}
	F=-E_{Q\left(\tilde{s},\pi\right)}\left[lnP\left(\tilde{o},\tilde{s},\pi| m\right)\right]-H\left[Q\left(\tilde{s},\pi\right)\right] \\[9pt]
	=E_{Q\left(\pi\right)}\left[-E_{Q\left(\tilde{s}|\pi\right)}\left[lnP\left(\tilde{o},\tilde{s}|\pi\right)\right]-H\left[Q\left(\tilde{s}|\pi\right)\right]\right]+KL\left[Q\left(\pi\right)| P\left(\pi\right)\right]\ \\[9pt]
	=E_{Q\left(\pi\right)}\left[F_\pi\right]+KL\left[Q\left(\pi\right)| P\left(\pi\right)\right]
	\end{gathered} \]
	
	Where $F_\pi$  is the energy of a policy over each time-step:
	\[ \begin{gathered}
	F_\pi=\ \sum_{\tau} F_{\pi \tau} 
	\\[12pt]
	F_{\pi\tau}=-E_{Q(s_{\tau}|\pi)Q(s_{\tau-1}|\pi)}[\left[\tau \le t\right]\cdot l n P\left(o_\tau | s_\tau\right)+lnP\left(s_\tau | s_{\tau-1},\pi\right)-lnQ\left(s_\tau |\pi\right)]
	\end{gathered}
	\]
	
	Having defined an objective function, beliefs about the hidden states may be iteratively optimised by gradient descent:  
	
	\[ \begin{gathered}
	{\dot{\hat{s}}}_\tau^\pi=\partial_{\hat{s}}\ s_\tau^\pi\  \cdot \varepsilon_\tau^\pi\ \\[7pt]
	s_{\tau}^\pi=\sigma\left({\hat{s}}_\tau^\pi\right) \\[7pt]
	\varepsilon_{\tau}^{\pi}=\ (\hat{A} \cdot o_{\tau}+ {\hat{B}}_{\tau-1}^\pi\ \cdot s_{\tau-1}^{\pi}+{\hat{B}}_\tau^\pi\ \cdot s_{\tau+1}^\pi)- \hat{s}_\tau^\pi  \\[5pt]
	=\  -\partial_sF
	\end{gathered} \]
	
	Solutions to the above equations converge toward posterior expectations that minimize free energy, providing Bayesian estimates of the hidden states that minimize prediction errors $\varepsilon_{\tau}^\pi$, expressed here as free energy gradients. This model encapsulates a single trial of the task lasting approximately 2 seconds, or a maximum of 8 saccades.
	
	The nature of the task requires the agent to utilise visual information that cannot be evaluated directly within the MDP. This issue is addressed by supplementing the agent’s generative model with an additional (subordinate) continuous-time model that can accumulate evidence from the visual domain \cite{friston2017graphical} i.e. directly from the attended pixels. In this model, conditional expectations $\tilde{\mu}$ about proprioceptive and exteroceptive sensory information are encoded by the internal states of the agent’s ‘brain’ in the form of a recognition density $q(x,v,a|\mu)$ that approximates the true posterior $ p(x,v,a|y,m) $, where $ y $ are the values of the attended pixels and the angular displacement of the eye. As before, this density can be optimised by maximizing Bayesian model evidence, or minimising variational energy:
	
	\[ F(\tilde{y},\tilde{\mu})=-lnp(y|m)+KL[q(x,v,a|\mu)||p(x,v,a|y,m)] \]
	
	By assuming a Gaussian form for the recognition density $p(x,v,a|y,m)$, we assume a local quadratic form for the variational free energy \cite{friston2008hierarchical} under the generative model:
	
	\[
	\begin{gathered}
	lnp\left(\tilde{y},\tilde{v},\tilde{x},\tilde{a}|\tilde{\mu}\right)=\frac{1}{2}ln\left|\tilde{\Pi}\right|-\frac{1}{2}{\tilde{\varepsilon}}^T\tilde{\Pi} \tilde{\varepsilon} \\[8pt]
	\tilde{\Pi}\ =\left[\begin{matrix}{\tilde{\Pi}}^v&&\\&{\tilde{\Pi}}^x&\\&&{\tilde{\Pi}}^a\\\end{matrix}\right] \\[9pt]
	\tilde{\varepsilon}=\left[\begin{matrix}{\tilde{\varepsilon}}^v=\ \left[\begin{matrix}y\\v^{\left(1\right)}\\\end{matrix}\right]-\left[\begin{matrix}{g}\\\eta\\\end{matrix}\right]\\[9pt]{\tilde{\varepsilon}}^x=\tilde{x}\ -\ f\\[7pt]\tilde{\varepsilon}^a=\ a\ -\ \eta\\\end{matrix}\right]
	\end{gathered}
	\]
	
	This formulation shows that the probabilistic generative model can be expressed in terms of prediction errors $ \tilde{\varepsilon} $ and their precision $ \Pi $. Estimates of the causal states $v$, their dynamics $x$ and action $a$ are derived from response $g(t)$ and state $ f(t) $ functions, described below. Empirical priors $ \eta  $ are descending messages that convey the expected outcomes from the \textit{what} and \textit{where} modalities of the superordinate MDP. 
	\begin{figure}[hbt!]
		\centering
		\includegraphics[width=13cm]{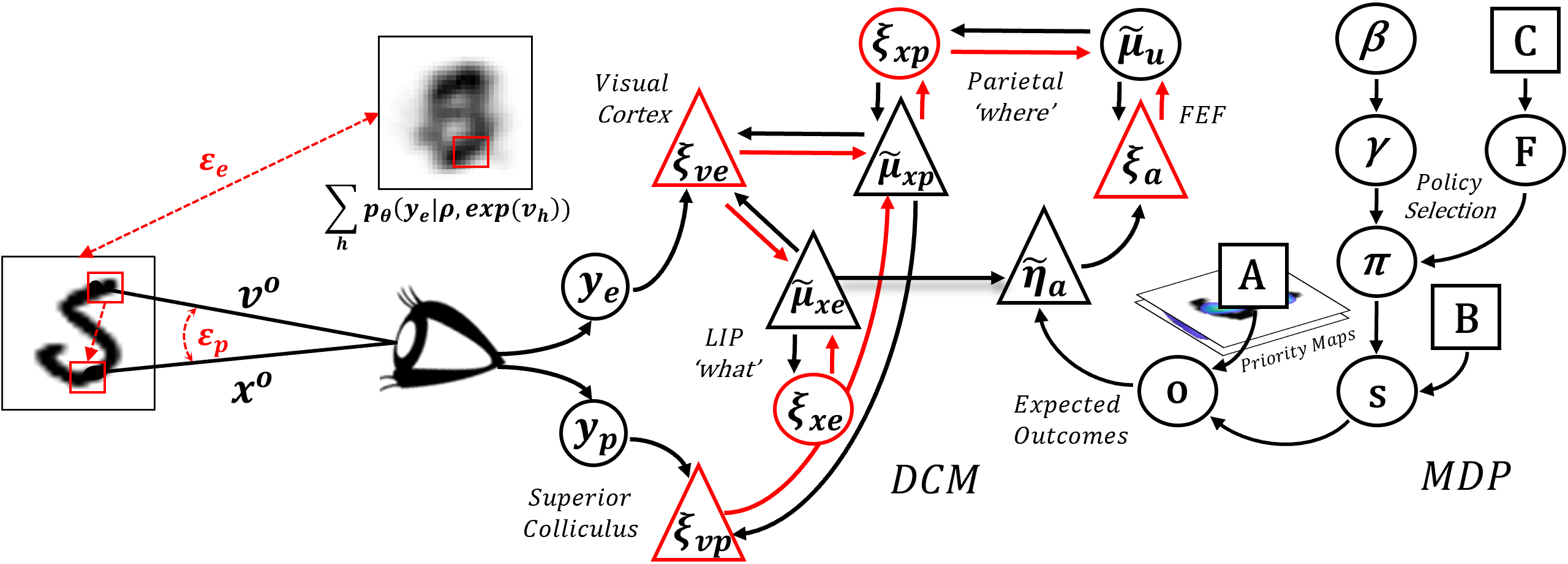}
		\caption{\textbf{Predictive Coding Scheme.}  By integrating this scheme targets of interest are selected and brought within the agent’s receptive field, inducing proprioceptive and exteroceptive stimuli $ y $ and prediction errors $ \varepsilon $. State-units encoding conditional expectations $ \tilde{\mu} $ are illustrated in black while error-units encoding precision-weighted prediction errors $ \xi $ are illustrated in red. The blurry prediction $ {\tilde{\mathbf{y}}}_\mathbf{e}\ $ on the left side of the image is generated from the prior network $ \mathbf{p}_\mathbf{\theta}\ $ under the weighted sum of random (uncertain) hypotheses $ \mathbf{v}_\mathbf{h} $. Exteroceptive prediction errors $ \mathbf{\varepsilon}_\mathbf{e} $ are derived from the absolute difference between this prediction and the observed stimulus at the sampling location $ \mathbf{x}^\mathbf{o} $. Proprioceptive prediction errors $ \mathbf{\varepsilon}_\mathbf{p} $ are derived from the difference between the current foveal location $ \mathbf{x}^\mathbf{o} $ and the saccade target $ \mathbf{v}^\mathbf{o} $ determined by top-down empirical priors $ \eta $ induced by policy selection in the MDP. As prediction errors are minimized, uncertainty is reduced and predictions become more accurate. MATLAB implementations of this optimisation scheme 'spm\_MDP\_VB\_X' and 'spm\_ADEM' are available to download as part of the SPM toolbox at \url{fil.ion.ucl.ac.uk/spm}}
	\end{figure}
	
	The agent interprets sensory information as though it were derived from two distinct modalities or streams; proprioceptive information $ y_{p} $ corresponds to the angular displacement of the eye, or the centre of gaze in extrinsic (cartesian) coordinates. Exteroceptive information $ y_{e} $ corresponds to visual stimuli sampled from a uniform grid $ \mathcal{R} $ of $ 8^{2} $ pixels centred around $ y_{p} $.
	
	\[ \begin{gathered}
	\tilde{y}\ =\ g =  \left[\begin{matrix}{\tilde{y}}_p\\{\tilde{y}}_e\\\end{matrix}\right]\ =\ \left[\begin{matrix}y_o\\R(y_e,y_o)\\\end{matrix}\right]+\omega \\[10pt]
	y_o= f = v_o-x_o \\[10pt]
	y_e=\sum_{h}{p_\theta\left(y_e|\rho,exp(v_h\right))}
	\end{gathered} \]
	
	Where the Gaussian innovations $ \omega $ induce small high-frequency perturbations (1-2 pixels) to the foveal sampling location. The causal states v comprise a 2D scalar target location $ v_{o} $ and digit class probabilities $ v_{h} $, both of which are prescribed by the superordinate MDP level. The final equality shows that competing visual hypotheses are scaled to reflect conditional uncertainty using the entropy of their (softmax) probabilities. The hidden states $x$ are the resulting motion of the eye relative to this location and the transitive values of the probability vector describing the subject’s belief about the target digit. 
	
	Posterior beliefs about the underlying causes of visual input $q_{\Theta}(z_{c},z_{d}|y) $ are optimised a-priori by a neural network with parameters $ \Theta $. This network encodes the sufficient statistics of a joint distribution over discrete $ z_{d}\equiv v_{h} $  and continuous $ z_{c}\equiv\rho $ factors of variation in the MNIST dataset. Categorical class probabilities $ z_{d} $ are encoded by Gumbel Softmax \cite{jang2016categorical} or concrete \cite{maddison2016concrete} distribution $ p_{\Theta}(z_{d}|y)=Gumbel(z_d) $. Variations in within-class visual features such as orientation and width are encoded by a multivariate normal distribution $ q_{\Theta}(z_{c_i}|y)\ =\ N(\mu_i,\sigma_i^2) $ with a unit prior $ p(z_c)=\mathcal{N}(0,1) $. The generative (prior) component of this model $ p_\theta(y| z_{c},z_{d}) $ is a neural network with parameters $ \theta $ that maps learned beliefs $ z_{c} $ and $ z_{d} $ to observations in the visual domain. Under the assumption that the discrete and continuous variables are conditionally independent, the objective function for both the posterior $ \Theta $ and prior $ \theta $ networks may be composed as per (Dupont, 2018) to facilitate the regularization of the discrete and continuous KL divergence terms during training:
	
	\[ \begin{aligned} L(\Theta,\theta)=\ E&_{q_\Theta(z_{c},z_{d}|y)}[logp_{\theta}(y| z_{c},z_{d})] \\[4pt]
	-& r_{c}|KL[{\ q}_{\Theta}(z_{c},y)\ ||\ p(z_{c})]-k_{c}| \\[4pt]
	-& r_{d}|KL[{\ q}_{\Theta}(z_{d},y)\ ||\ p(z_{d})]-k_{d}|
	\end{aligned} \]
	
	Where $r$  and $k$ are free regularisation parameters. Doing so encourages disentanglement \cite{higgins2017beta} of the latent variables, allowing for each factor of variation in the data to be learned and subsequently coupled with one or more outcome modalities in the MDP. Here we are concerned only with the disentanglement of the target classes and their relationship to the $ \mathit{digit} $ outcome modality, but this technique may prove useful for other experimental paradigms. With this model we may describe the dynamics of the task, including semantic content and ocular-motor dynamics as differential equations that are integrated with respect to sensory input over the duration of a single saccade (~200ms), or a single state transition within the Markov process. 
	
	Communication between the Markov process and the dynamic model is mediated by a link function that transforms ascending prediction error messages ${\tilde{\varepsilon}}^v $ from the continuous-time model into posterior expectations $\tilde{o}$ over outcomes in the Markov model and descending predictions from the Markov model into empirical priors $ \eta $ at the continuous level.
	
	We define a set of reduced (competing) models $ \vartheta $ by collapsing a prior density over each possible outcome $R\in[1,10x50]$. Each reduced model $ \vartheta_{m} $ is a prior encoding a visual hypothesis at a target saccade location, evaluated over the duration of the saccade: 
	
	\[ 
	\begin{aligned}
	{E\left(t\right)}_m=&-lno_{\tau,m}-\int_{0}^{T}{{L\left(t\right)}_mdt} \\[3pt]
	{L\left(t\right)}_m=&lnP\left(\tilde{y}\left(t\right)|\vartheta_m\right)-lnP(\tilde{y}_{t}|\eta)
	\end{aligned}
	\]
	\[
	\\[4pt]
	\begin{array}{cc}
	o_\tau=\sum \pi_\pi\cdot o_\pi^\tau  & \vartheta=\sum\vartheta_m\cdot o_m^\tau 
	\end{array} 
	\]
	
	The free energy E at the last time-step of the sequence takes the place of posterior expectations over outcomes $ o $ in the MDP. See \cite{friston2017graphical} for a neurobiological interpretation of this function.
	
	\section{Results}
	We first demonstrate that our model can categorize digits based on limited foveated sampling. In this example we see that the model builds its beliefs about the identity of the digit in only three saccades (Figure 3). 
	
	\begin{figure}[hbt!]
		\centering 
		\includegraphics[width=11cm]{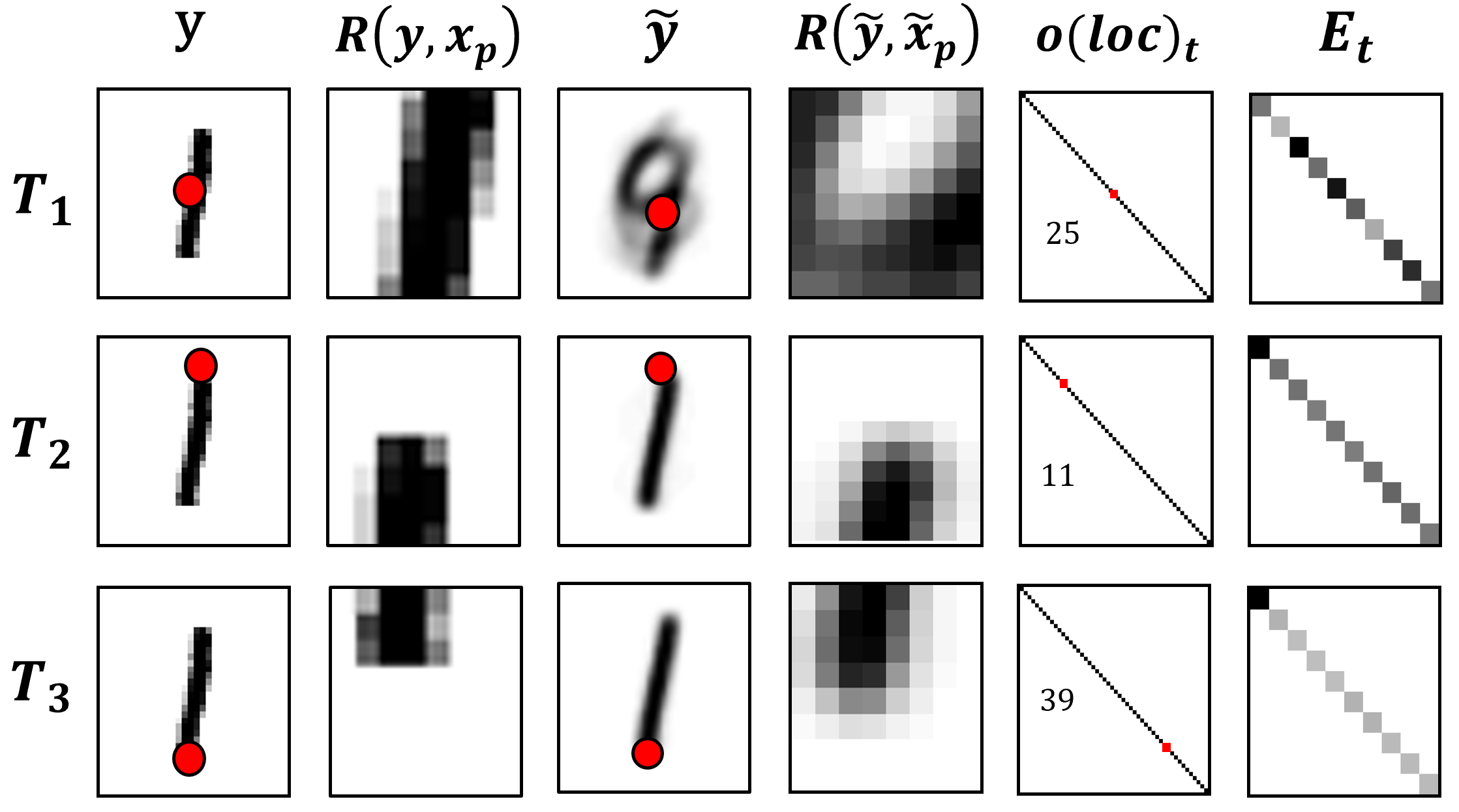}
		\caption{\textbf{Model Outcomes}. Column 1) Unfoveated task stimulus $ y $, with a red dot marking the foveation location (which neither the model of human participant could see). Column 2) Task stimulus foveated around $ x_{p} $ (observable by model and human participants). Column 3) The agent’s visual expectation about the global scene generated from the prior network $ p_{\theta} $. Column 4) The agent’s visual expectation about the stimulus at $x_{p} $. 5)  Posterior expectation about the target location $ o_{l} $. 6) Ascending posterior expectation about the target digit $ o_{h} $.}
	\end{figure}
	
	To estimate subject-specific parameters from observed behaviour we specify an objective model $ m^{o} $ in terms of the likelihood $ p (y |\theta,\lambda(\vartheta),u,m^{o}) $ of (the participant’s) behavioural responses $ y $ and a prior over the unknown parameters $ p(\vartheta,\theta|m^{o}) $. The implicit generative model $ \lambda(\vartheta) $, with subjective parameters $ \vartheta\in x,v,\rho $, in the likelihood function provides a differentiable mapping from task stimuli $u$ to participant’s behaviour $y$. The unknown parameters of the objective model $ \theta\in C,\beta $ correspond to putative neurobiological quantities that we wish to infer; we focus here on the precision of prior preferences over outcomes C, which we presume to be encoded by the ventromedial prefrontal cortex \cite{paulus2003ventromedial} and the inverse precision of beliefs about control states $ \beta $, which we presume to be encoded by dopaminergic projections from ventral tegmental area and substantia nigra to the striatum \cite{fitzgerald2015dopamine}\cite{schwartenbeck2015dopaminergic}.
	
	We optimise this objective model to recover estimates of $C$ and $\beta$ with respect to observed behaviour (Figure 4) using the same variational technique as the subjective model i.e gradient descent on variational free energy \cite{daunizeau2009variational}.
	
	\begin{figure}[hbt!]
		\centering
		\includegraphics[width=13cm]{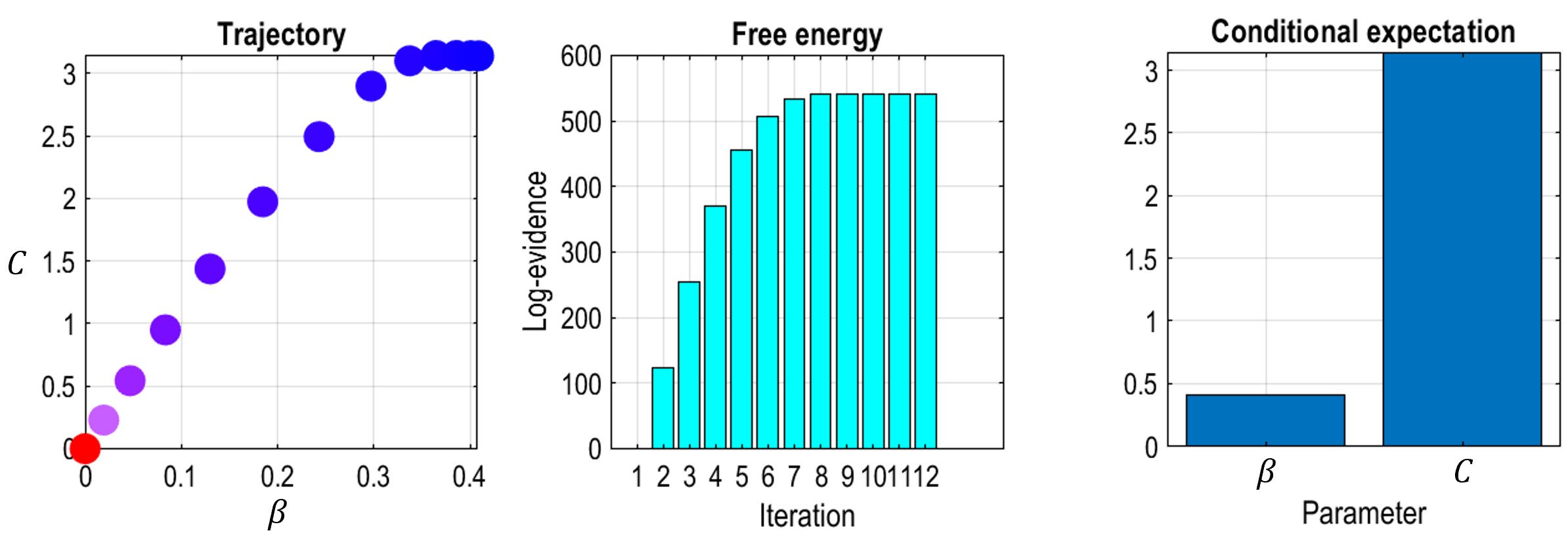}
		\caption{\textbf{Model inversion with respect to observed behaviour}. A) The trajectory of the estimated parameters in parameter space. B) The lower bound of the log-model evidence approximated as variational free energy. C) Final conditional parameter estimates.}
	\end{figure}
	\begin{figure}[hbt!]
		\centering
		\includegraphics[width=13cm]{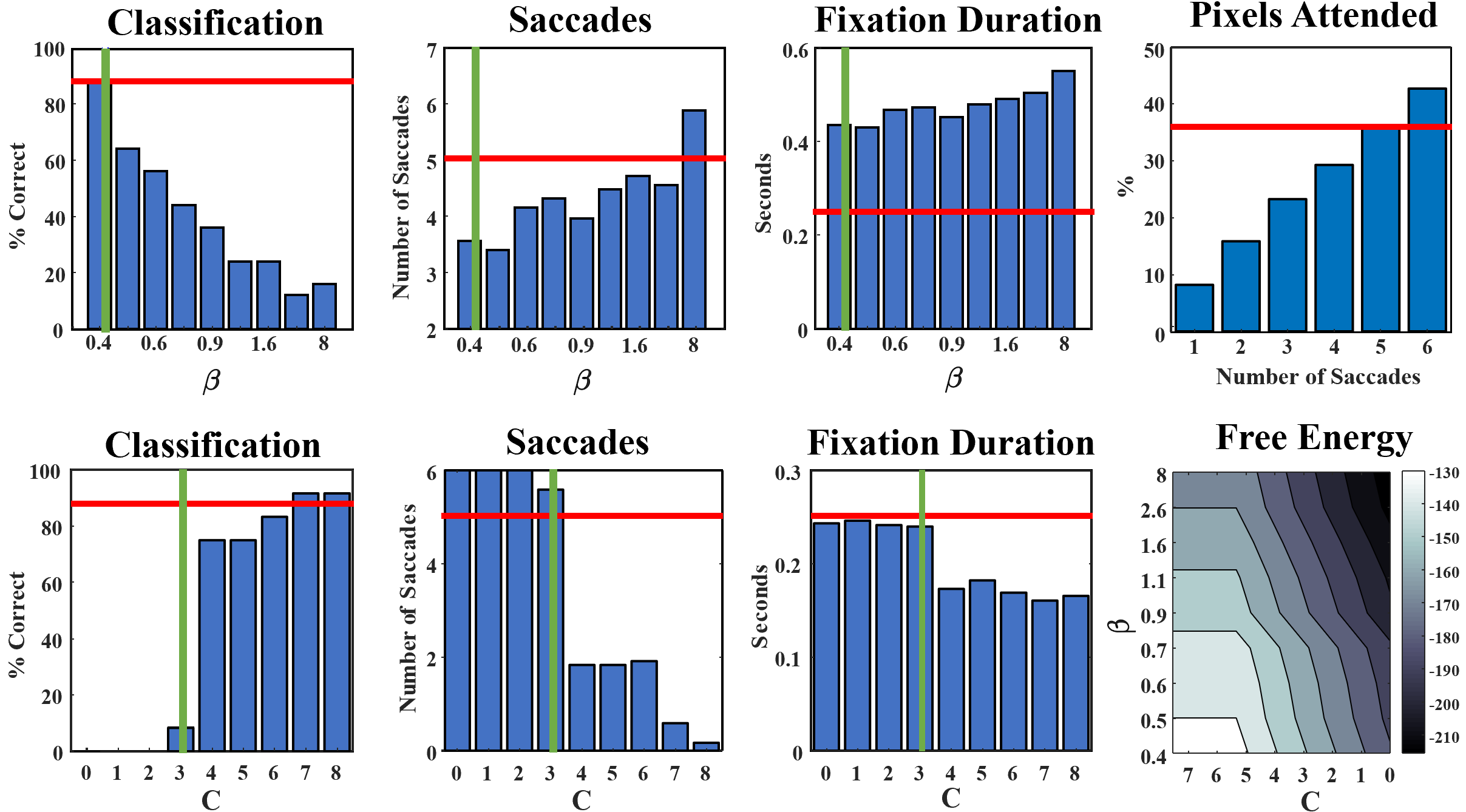}
		\caption{\textbf{Behavioural metrics as a function of model parameters.} Red (horizontal) lines indicate the mean human responses. Green (vertical) lines indicate the parameter values recovered from model inversion. The top row displays these metrics as a function of policy precision $ \gamma $ with inverse temperature $ \beta $. As $ \beta $ increases, or as the agent's confidence in its actions decreases, accuracy declines, fixations become longer and the total number of saccades increase. The top-right figure shows the average percentage of (unique) attended pixels as a function of the total number of saccades. The bottom row shows that the accuracy of the model increases as it’s intrinsic motivation to observe \textit{correct} feedback (C) increases, and that for C<=3, the model is not incentivised to make a decision. The bottom-right figure shows the average free energy of the model as a function of the model parameters. }
	\end{figure}
	
	By simulating 100 trials for each digit class with the parameters recovered from model inversion, we find that on average, the correct digit is inferred on 88\% of trials after 5.02 saccades (Figure 5). This corresponds to 35.7\% pixels viewed relative to the total number of pixels in the image. The code used to generate these results is available to download from \url{https://github.com/v2c08/M-BMVS}.
	
	\section{Discussion}
	An important mechanistic assumption made by the Free Energy Principle and other related Bayesian Brain theories \cite{doya2007bayesian}\cite{knill2004bayesian} is that the physical states of sensory systems encode probabilistic representations of the environment that interact over multiple spatial and temporal scales. Empirical support for these theories comes primarily from studies of predictive coding \cite{friston2009predictive} in the sensory cortices \cite{bastos2012canonical}\cite{shipp2016neural}\cite{spratling2010predictive}. Theories of predictive coding cast the cerebral cortex as a Bayesian generative model that implements approximate (variational) inference through a recursive interchange of prediction and error signals. We appeal to this theory in our work and present a hierarchical model of visual search that incorporates Bayesian interpretations of the physiological processes governing perception and behaviour. 
	
	While the model may seem complex, its form and the attendant variational optimisation scheme can be generalised to most if not all behavioural paradigms. The discrete Markov scheme has previously been used to model context learning \cite{friston2016active}, goal-directed behaviour \cite{pezzulo2018hierarchical}, addiction \cite{schwartenbeck2015optimal}, scene construction \cite{mirza2016scene}, and reading \cite{friston2018deep} while the dynamic scheme has been used to simulate perception \cite{kiebel2009perception}, attention \cite{feldman2010attention} and sensorimotor integration \cite{friston2011optimal}. 
	
	However, in many cases the coupled differential equations that define the response and dynamic functions cannot be specified due to the complexity of the task-relevant data, i.e. high-dimensional inputs with causal or temporal relationships between modalities. In this work we demonstrate the use of unsupervised generative neural networks in place of the unknown or intractable response function. Appealing once again to predictive coding, we draw a comparison  between learned visual representations in the brain and the latent variables of the variational autoencoder. We presume that representations in the brain are derived from prediction errors invoked by sensory interactions with the environment, and that the compositional structure of these representations reflect the causal structure of the stimuli from which they are derived. The scheme presented here is particularly well suited to tasks in which representations of task-relevant stimulus categories can be ‘disentangled’ by neural networks. Doing so allows the experimenter to account for each factor of variation in the experimental design, and their latent representations in the subjective ‘forward’ model of the task. 
	
	We have identified several ways in which our model may be extended, for example to facilitate the categorisation of naturalistic images; to do so one may consider replacing our relatively simple decoder network $ p_{\theta} $ with a more powerful autoregressive decoder such as \cite{gulrajani2016pixelvae}. It may also be possible to integrate other (interpretable) mechanisms for class-contingent salience attribution such-as \cite{elsayed2019saccader}, allowing the agent to evaluate the likelihood of policies defined over internal representations of space, rather than a simple 2D grid.  One may also wish to improve the proprioceptive component of our model by incorporating continuous ocular kinetic parameters within the subjective generative model as per \cite{adams2015active}. Doing so would allow the model to more accurately account for individual differences in occular-motor dynamics. 
	
	\section{Experimental Procedures}
	A visual eye-tracking study was performed on healthy participants to analyse scan paths during visual search. Movements of the left eye were tracked and recorded at 1kHz with an Eyelink1000 (SR Research) system. A 12-point calibration procedure was implemented at the beginning of each recording session. We used a Windows 7 desktop computer and a monitor displaying at a resolution of 1280 x 1024 @ 85.3 Hz. This experiment was realised using Cogent Graphics developed by John Romaya at the LON at the Wellcome Department of Imaging Neuroscience, and Psychophysics Toolbox extensions for MATLAB \cite{brainard1997psychophysics}\cite{kleiner2007s}. In total 28 participants were recruited, aged 20-34 (M=26.5, SD=3.5), all participants had a visual acuity of 20/20 read from a Pocket Snellen chart. Approval was granted by the KCL Research Ethics Subcommittee Ref:MRM-18/19-11544 and all participants gave written informed consent.
	
	\newpage
	%
	%
	
	\bibliography{mbmvs_arxiv}
\end{document}